\documentclass[letterpaper]{article} % DO NOT CHANGE THIS
\usepackage{aaai2026}  % DO NOT CHANGE THIS

\usepackage{times}  % DO NOT CHANGE THIS
\usepackage{helvet}  % DO NOT CHANGE THIS
\usepackage{courier}  % DO NOT CHANGE THIS
\usepackage[hyphens]{url}  % DO NOT CHANGE THIS
\usepackage{graphicx} % DO NOT CHANGE THIS
\usepackage{natbib}  % DO NOT CHANGE THIS AND DO NOT ADD ANY OPTIONS TO IT
\usepackage{caption} % DO NOT CHANGE THIS AND DO NOT ADD ANY OPTIONS TO IT
\usepackage{algorithm}
\usepackage{algorithmic}

\usepackage{newfloat}
\usepackage{listings}
\DeclareCaptionStyle{ruled}{labelfont=normalfont,labelsep=colon,strut=off} % DO NOT CHANGE THIS
\floatstyle{ruled}
\newfloat{listing}{tb}{lst}{}
\floatname{listing}{Listing}
\usepackage{booktabs}
\usepackage{amsmath}
\usepackage{amsfonts}
\usepackage{multirow}

\title{ReGal: A First Look at PPO-based Legal AI for Judgment Prediction and Summarization in India}
\author{
    Shubham Kumar Nigam\textsuperscript{\rm 1, 5}\equalcontrib\thanks{Corresponding Author},
    Tanuj Tyagi\textsuperscript{\rm 2}\equalcontrib,
    Siddharth Shukla\textsuperscript{\rm 2}\equalcontrib,
    Aditya Kumar Guru\textsuperscript{\rm 2}\equalcontrib,
    Balaramamahanthi Deepak Patnaik\textsuperscript{\rm1 }\equalcontrib,
    Danush Khanna\textsuperscript{\rm 2}\equalcontrib,
    Noel Shallum\textsuperscript{\rm 3}\equalcontrib, 
    Kripabandhu Ghosh\textsuperscript{\rm 4},
    Arnab Bhattacharya\textsuperscript{\rm 1}
}
\affiliations{
     \textsuperscript{\rm 1}Indian Institute of Technology Kanpur, India \quad
     \textsuperscript{\rm 2}Manipal University Jaipur, India \quad
     \textsuperscript{\rm 3}Symbiosis Law School Pune, India \quad
     \textsuperscript{\rm 4}IISER Kolkata, India \quad
     \textsuperscript{\rm 5}University of Birmingham, Dubai, United Arab Emirates\\
    \texttt{\{shubhamkumarnigam, tanujtyagiofficial, siddharth8shukla8, adityaguru20, bdeepakpatnaik2002, danush.s.khanna, noelshallum\}@gmail.com} \\ 
\quad \texttt{kripaghosh@iiserkol.ac.in} \quad \texttt{arnabb@cse.iitk.ac.in}
}

\usepackage{bibentry}
\begin{document}

\maketitle

\begin{abstract}
    This paper presents an early exploration of reinforcement learning methodologies for legal AI in the Indian context. We introduce \textsc{Reinforcement Learning-based Legal Reasoning} (\texttt{ReGal}), a framework that integrates Multi-Task Instruction Tuning with Reinforcement Learning from AI Feedback (RLAIF) using Proximal Policy Optimization (PPO). Our approach is evaluated across two critical legal tasks: (i) Court Judgment Prediction and Explanation (CJPE), and (ii) Legal Document Summarization. Although the framework underperforms on standard evaluation metrics compared to supervised and proprietary models, it provides valuable insights into the challenges of applying RL to legal texts. These challenges include reward model alignment, legal language complexity, and domain-specific adaptation. Through empirical and qualitative analysis, we demonstrate how RL can be repurposed for high-stakes, long-document tasks in law. Our findings establish a foundation for future work on optimizing legal reasoning pipelines using reinforcement learning, with broader implications for building interpretable and adaptive legal AI systems.
\end{abstract}
\begin{links}
    \link{Code}{https://github.com/ShubhamKumarNigam/ReGal}
\end{links}


% Keywords: Artificial Intelligence, Judicial Decision Support, Instruction Tuning, Reinforcement Learning, Indian Legal System, Judgment Prediction and Explanation, Reinforcement Learning from AI Feedback (RLAIF), Legal Reasoning, Explainable AI (XAI)
% \end{abstract}

\section{Introduction}
This paper explores the use of Reinforcement Learning (RL) for two crucial tasks in Indian legal AI: Court Judgment Prediction and Explanation (CJPE) and Legal Document Summarization. While recent works have explored various approaches for judgment prediction \cite{ILDC2021, vats-etal-2023-llms, nigam-etal-2024-legal} and rhetorical segmentation \cite{bhattacharya2019comparative, malik2021semantic}, the integration of reinforcement learning, especially using Proximal Policy Optimization (PPO), remains largely unexplored in this domain. Our work marks one of the first attempts to apply PPO-based RL techniques to Indian legal tasks, addressing the unique challenges posed by the Indian judiciary system.

The CJPE task involves two interlinked subtasks: prediction of the case outcome and generation of a rationale explanation based on factual case records. While prior research in this area has primarily relied on supervised fine-tuning of pretrained language models \cite{nigam2023fact, nigam2023nonet, katz2017general, zhu2020legal}, such approaches are often limited by their reliance on large, annotated datasets and their inability to incorporate real-time feedback for interpretability. Our framework, \textsc{Reinforcement Learning-based Legal Reasoning} (\texttt{ReGal}), addresses this gap by employing RL to iteratively refine both decisions and explanations using a reward signal derived from legal textual alignment.

In addition to CJPE, we also extend our framework to the legal summarization task, which is a critical tool for improving access to justice and aiding practitioners in quickly understanding long judgments. We experiment on the In-Abs summarization dataset, a benchmark curated for generating abstract-style summaries from Indian court decisions. While summarization has seen advances through neural and transformer-based models \cite{summarizationLLMIndia, shukla2022legal, datta-etal-2023-mildsum, joshi-etal-2024-il}, the application of RL-based optimization in this setting, especially under Indian law, remains underexplored. By testing our PPO-based approach on both CJPE and summarization, we demonstrate that reinforcement learning has cross-task potential in legal NLP, despite facing several challenges in terms of performance and hallucination.

While \texttt{ReGal} underperforms compared to fine-tuned and proprietary models like GPT-3.5, our findings reveal critical insights into why RL techniques struggle with legal texts, including reward model misalignment, domain complexity, and lack of domain-adaptive pretraining. The ablation studies, hallucination examples, and comparative analysis on lexical and semantic metrics collectively highlight the limitations of current RLHF techniques in handling nuanced, high-stakes legal language.

This work should be viewed as a position paper that lays the groundwork for future improvements. Rather than focusing solely on state-of-the-art performance, we emphasize methodological exploration, share valuable lessons, and propose actionable directions for advancing RL in legal AI.

% To ensure reproducibility, we release the code and data through an anonymous link\footnote{\url{https://github.com/ShubhamKumarNigam/ReGal}}.
Our contributions are threefold: (1) we present one of the first applications of PPO-based reinforcement learning in Indian legal judgment prediction and summarization; (2) we provide empirical and qualitative evidence on its limitations; and (3) we chart a path for more effective legal-AI pipelines integrating RLHF, human feedback, and domain-adapted modeling.

\section{Related Work}
\textbf{Legal Judgment Prediction (LJP)} Legal judgment prediction has been explored across various jurisdictions using SVMs, CNNs, transformers, and now LLMs \citep{aletras2016predicting, chalkidis2019neural, feng2021recommending}. Benchmarks like CAIL2018 \citep{xiao2018cail2018}, ECHR \citep{chalkidis2019neural}, and TopJudge established task formulations in Chinese and European legal systems.

In India, ILDC \citep{malik-etal-2021-ildc}, PredEx \citep{nigam-etal-2024-legal}, and NyayaAnumana \citep{nigam-etal-2025-nyayaanumana} provided datasets for factual and explainable LJP. \citet{nigam2023fact, nigam2023nonet} proposed hierarchical transformers and fact-only predictions. Other studies \citep{kapoor-etal-2022-hldc, ganguly2023legal} focused on Hindi documents and long-text summarization.

LJP has also been studied cross-jurisdictionally \citep{zhao-etal-2018-learning}, in Romania \citep{masala2021jurbert}, Korea \citep{hwang2022multi}, and Switzerland \citep{niklaus2021swiss}. However, no prior work has used RLHF or RLAIF to refine predictions in Indian courts.

\textbf{Legal Judgment Summarization} Legal case summarization is challenging due to the length and rhetorical structure of court documents. Extractive and abstractive summarization using models like BART, PEGASUS, and LED have been explored \citep{shukla-etal-2022-legal, feijo2023improving}. Older rule-based and graphical models like CaseSummarizer \citep{polsley-etal-2016-casesummarizer} and DelSumm \citep{bhattacharya2021incorporating} have also been proposed.

In the Indian context, most legal summarization work remains extractive. No prior work has utilized reinforcement learning for Indian legal summarization.

\textbf{Reinforcement Learning for Legal Summarization} Several prior works have explored the use of reinforcement learning in the legal domain, particularly for tasks such as legal summarization. For instance, \citet{shukla2022text} investigated the application of policy gradient methods to improve summarization quality by aligning model outputs with reward signals derived from human feedback. Similarly, \citet{wang2024empowering} proposed a novel RL framework based on Soft Actor-Critic with Variational Autoencoders (SAC-VAE), demonstrating its effectiveness in generating high-quality legal summaries. In other lines of work, \citet{nguyen2021robust}, \citet{dong2024reinforcement}, and \citet{patil2025reinforcement} introduced value-based methods such as Deep Q-Networks (DQN) and Advantage Actor-Critic (A2C) models to reinforce generation strategies in legal texts, enabling more contextually grounded and fluent outputs. These studies collectively highlight the growing interest in using RL techniques to align model behavior with human expectations in legal NLP applications.

However, these remain extractive or domain-agnostic. Our work is the first to explore abstractive legal summarization in Indian courts using RLHF and RLAIF.

\textbf{Instruction Tuning and RLAIF} Instruction tuning (IT) helps align LLMs with user intents \citep{wei2022finetuned, mishra2022crosstask, zhou2023-InstructCTG}. Scaling IT with diverse prompts improves controllability and task generalization \citep{zhang2023instruction, wang2022supernaturalinstructions}.

RLHF \citep{ouyang2022-instuctGPT, glaese2022improving} improves LLM alignment with human preferences. RLAIF \citep{lee2023rlaif} offers a cost-effective alternative using AI-generated feedback, achieving near-RLHF performance \citep{ziegler2020finetuning, bai2022constitutional}.

To our knowledge, this is the first application of RLAIF and RLHF for both LJP and summarization in the Indian legal domain.

\textbf{LLMs for Summarization and Other Domains} LLMs like GPT-4 and Claude have shown strong summarization abilities for long books \citep{chang2023booookscore}, news articles \citep{zhang2023benchmarking}, and meeting transcripts \citep{schneider2023team}.

Hierarchical summarization \citep{chang2023booookscore} and zero-shot/few-shot prompting have demonstrated LLMs' potential in generating abstractive summaries. Our work extends these paradigms to Indian legal documents. While datasets like CrossSum \citep{bhattacharjee2021crosssum}, WikiLingua \citep{ladhak-etal-2020-wikilingua}, and ILSUM \citep{urlana2023indian} cover Indian languages, they lack legal domain coverage.

Eur-Lex \citep{aumiller-etal-2022-eur} and CLIDSUM \citep{wang-etal-2022-clidsum} offer multilingual legal summaries, but no such benchmark exists for Indian court judgments. Our work bridges this gap via expert-annotated data and RL tuning.

To the best of our knowledge, this is the first work that applies RLHF/RLAIF to both legal judgment prediction and summarization tasks in the Indian legal context.

% Prior research in Court Judgment Prediction (CJP) has explored both international and Indian jurisdictions using machine learning and deep learning approaches \citep{aletras2016predicting, chalkidis2019neural, feng2021recommending, nigam-etal-2024-legal}. Notable efforts in India include ILDC \citep{malik-etal-2021-ildc} and PredEx \citep{nigam-etal-2024-legal}, which emphasize factual and explainable AI.

% Summarization of legal documents has been attempted using extractive and abstractive methods \citep{shukla-etal-2022-legal, polsley-etal-2016-casesummarizer}, and more recently, LLMs \citep{zhang2023benchmarking, chang2023booookscore}. However, limited work exists on Indian legal judgment summarization, especially using reinforcement learning.

% Instruction tuning and Reinforcement Learning from Human or AI Feedback (RLHF, RLAIF) have shown success in aligning LLM outputs with human preferences across domains \citep{ouyang2022-instuctGPT, lee2023rlaif}. While these methods have been used in tasks like summarization and QA, their application in the Indian legal domain has not been explored before.

\section{Task Description}

This paper presents the ReGal framework, which combines instruction tuning with Reinforcement Learning from AI Feedback (RLAIF) to enhance legal NLP systems. As a position paper, our goal is to evaluate the general applicability of this architecture across multiple tasks in the Indian legal domain, specifically focusing on reinforcement learning's capacity to refine predictions and enhance interpretability.

\begin{figure*}[t] 
    \centering 
    \includegraphics[width=0.55\linewidth]{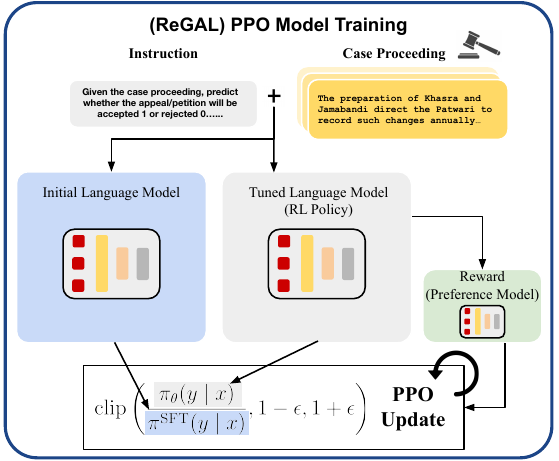} 
    \caption{Overview of the \texttt{ReGal} PPO model training process.}
    \label{fig:task-framework}
\end{figure*}

Figure~\ref{fig:task-framework} illustrates the overall architecture and PPO training pipeline used across tasks. Rather than being specific to a single task, this figure highlights how the same optimization loop, grounded in expert or model-generated feedback, is flexibly applied to different legal reasoning tasks. While our experiments are conducted on Indian Supreme Court judgments, the ReGal framework is task-agnostic and potentially extensible to other domains requiring high interpretability and reasoning fidelity.

We evaluate ReGal on the following two core tasks:

\subsection*{Task 1: Court Judgment Prediction and Explanation (CJPE)}
This task assesses the model’s ability to reason over complex legal documents and consists of two tightly coupled subtasks:

\paragraph{Task 1A:  Judgment Prediction}  
Given a legal document \( D \), typically a judgment from the Supreme Court of India, the goal is to predict whether the appeal or petition was accepted or rejected, represented as binary labels \( y \in \{0, 1\} \). This task reflects the practical need for predictive tools in legal practice and case screening.

\paragraph{Task 1B:  Rationale Explanation}  
In this subtask, the model is expected to generate a natural language explanation supporting its predicted outcome. The explanation should reflect the key reasoning patterns within the case text and ideally mimic judicial argumentation. This component enhances the interpretability and trustworthiness of the AI system.

\subsection*{Task 2: Legal Judgment Summarization}
To demonstrate the generalization of the ReGal framework, we also evaluate it on the task of judgment summarization. Here, the model must generate a concise yet informative summary of the full judgment text, capturing its essential components such as background, legal issues, arguments, and final verdict. This task reflects a broader class of document understanding problems and allows us to assess the effect of reinforcement learning in improving content selection and abstraction in long legal texts.

By applying a unified PPO-based training regime across these tasks, we argue that ReGal provides a promising and extensible approach for reinforcement-tuned legal AI. The tasks differ in their output formats and supervision requirements, but share a common challenge: the need for factual consistency, interpretability, and domain alignment, properties that our reinforcement learning strategy seeks to enhance.

%\section{Task Description}

% This work aims to advance the Court Judgment Prediction and Explanation (CJPE) task. Figure \ref{fig:task-framework} provides a visual overview of the CJPE task framework employed in this study, highlighting the sequential nature of the prediction and explanation processes. The figure also emphasizes the role of Large Language Models (LLMs) and the Reinforcement Learning from Human Feedback (RLHF) pipeline in our approach, showcasing how these technologies are integrated into the task workflow.

% The CJPE task is broken down into two subtasks:

% \paragraph{Task A: Judgment Prediction Task:}
% The primary goal of this subtask is to predict the outcome of a legal case based on the provided case proceedings. Given a document $D$ containing case proceedings from the Supreme Court of India (SCI), the objective is to predict the decision \(y \in \{0, 1\}\), where 1 represents the acceptance of the appeal or petition by the appellant or petitioner, and 0 denotes its rejection.

% \paragraph{Task B: Rationale Explanation Task:}
% In this subtask, the goal is to generate a coherent explanation or rationale that justifies the predicted decision. This explanation is based on the relevant segments of the judgment, providing clarity and insight into the reasoning behind the predicted outcome. By ensuring that the rationale is well-articulated, we aim to enhance the transparency and interpretability of the prediction process, ultimately contributing to the broader acceptance of AI-driven tools in the legal domain.

%%%%%%%%%%%%%%%%%%%%%%%%%%%%%%%%%%%%
%%%%%%%%%%%%%%%%%%%%%%%%%%%%%%%%%%%%

\section{Dataset}

To evaluate the proposed ReGal framework across diverse legal tasks, we utilized two large-scale, expert-curated datasets focused on reasoning and summarization in the Indian legal domain.

\paragraph{CJPE Task Dataset — PredEx:}
For the Court Judgment Prediction and Explanation (CJPE) task, we use the publicly available PredEx dataset introduced by~\citep{nigam-etal-2024-legal}. It is the largest annotated dataset in the Indian legal NLP space for judgment prediction and rationale generation. Developed with inputs from ten senior law students across reputed institutions, the dataset ensures annotation quality through expert-in-the-loop processes conducted between April 2022 and October 2023.

PredEx contains 15,222 Supreme Court judgment documents, split into training (12,178) and test (3,044) sets. Each document includes the full case text, a binary outcome label (accepted/rejected), and a corresponding human-written explanation for the decision. Summary statistics are provided in Table~\ref{data-stats}.

\begin{table}[t]
  \centering
\resizebox{0.9\columnwidth}{!}{%
  \begin{tabular}{lrr}
    \toprule
    \textbf{Metric} & \textbf{Train} & \textbf{Test} \\
    \midrule
    No. of documents & 12,178 & 3,044 \\
    Average no. of tokens & 4,586 & 4,422 \\
    Minimum no. of tokens & 176 & 184 \\
    Maximum no. of tokens & 117,733 & 83,657 \\
    Acceptance percentage & 53.44 & 50.00 \\
    \bottomrule
  \end{tabular}
}
    \caption{Statistics of the PredEx dataset~\citep{nigam-etal-2024-legal} used for judgment prediction and explanation.}
  \label{data-stats}
\end{table}

\paragraph{Summarization Task Dataset — In-Abs:}
For the legal document summarization task, we employ the In-Abs subset of the IL-TUR dataset~\citep{joshi-etal-2024-il,shukla2022legal}. This dataset includes 7,130 Supreme Court documents paired with expert-written headnotes, which serve as abstractive summaries capturing critical aspects such as legal issues, arguments, and verdicts. These summaries, written by legal experts and extracted using heuristic rules, provide high-quality ground truth for evaluating summarization quality.

The dataset is split into 7,030 training and 100 test samples. Table~\ref{tab:inabs-stats} shows key statistics.

\begin{table}[t]
\centering
\begin{tabular}{ll}
\toprule
\textbf{Dataset} & \textbf{In-Abs} \\
\midrule
\# Documents & 7,130 \\
Type of Summary & Abstractive \\
Language & English \\
Train/Test Split & 7,030/100 \\
Avg. Document size (in \#words) & 4376.98 \\
Avg. Summary size (in \#words) & 842.52 \\
Avg. Compression Ratio & 0.235 \\
\bottomrule
\end{tabular}
\caption{Statistics of the In-Abs summarization dataset~\cite{shukla2022legal}.}
\label{tab:inabs-stats}
\end{table}

Both datasets allow us to assess the applicability of reinforcement learning via PPO across diverse generative tasks, prediction, explanation, and summarization, within Indian legal NLP, demonstrating the flexibility of our approach.

\section{Methodology}
We propose a reinforcement learning framework, \texttt{ReGal}, that aims to enhance large language models (LLMs) for legal reasoning tasks in the Indian judicial domain. Our framework is instantiated over two distinct legal tasks: (i) Court Judgment Prediction and Explanation (CJPE), and (ii) Legal Summarization. For both tasks, we adopt a two-stage approach involving supervised fine-tuning followed by reinforcement learning via Proximal Policy Optimization (PPO), using AI-generated reward signals. This setup enables us to explore the applicability and effectiveness of PPO across multiple legal generation objectives.

\subsection{Base Model and SFT Training}
For all tasks, we adopt Llama-2-7B as the base language model. This decision is motivated by its prior use in recent literature on legal judgment prediction and explanation, particularly within the Indian legal context~\cite{nigam-etal-2024-legal}. By choosing the same model, we enable a fair comparison between our proposed reinforcement learning-based alignment (ReGal) and earlier fine-tuning approaches. This consistency also helps isolate the effects of our PPO optimization, allowing us to demonstrate the specific contribution of reinforcement learning to performance gains.

We fine-tune Llama-2-7B using supervised instruction tuning on both tasks: for CJPE, this includes predicting the case outcome and generating explanations using the PredEx dataset; and for Summarization, the model is trained to generate abstractive summaries using the IL-TUR dataset. This supervised fine-tuned model, denoted as \( \pi^{SFT} \), serves as the reference policy in subsequent reinforcement learning stages.

% \subsection{Reward Model}
% To enhance the performance of the base model, we employ a fine-tuned reward model (RM) based on InLegalBERT. This reward model is designed to classify judgments as accepted or rejected, providing a feedback mechanism that aligns with our prediction objectives. During the training process, we assign a reward of 1 for accurate predictions and a reward of 0 for incorrect ones. This reward structure incentivizes the model to generate outputs that closely align with our goals for prediction and explanation. The reward model acts as a proxy for human evaluation, allowing the aligned language model to learn from AI-generated feedback.

\subsection{Reward Models}

To align the outputs with the desired objectives, we develop task-specific reward models (RMs). For CJPE, we fine-tune InLegalBERT to classify the correctness of predictions. The RM assigns binary rewards: 1 if the model predicts the correct verdict, and 0 otherwise. For Summarization, the RM is trained to score summaries based on n-gram overlap with the gold headnotes and coherence. We use ROUGE-style matching and shallow semantic similarity to assign scalar rewards.

These RMs simulate AI-based feedback in lieu of human-in-the-loop supervision.

\subsection{Proximal Policy Optimization (PPO)}
To address the objective mismatch in the fine-tuned model and improve the alignment between predicted outputs and desired outcomes, we employ Proximal Policy Optimization (PPO). PPO is a widely used actor-critic algorithm designed to optimize policies in reinforcement learning, especially for large language models. It allows us to adjust the model's behavior in response to feedback by iteratively updating its parameters, thus aligning the model more effectively with the prediction and explanation tasks.

In this context, we treat the aligned language model as a learnable policy \( \pi_{\theta} \), where \( \theta \) represents the model's parameters. PPO optimizes this policy to minimize the discrepancy between the model’s predicted outputs and the expected outcomes, as dictated by a reward function. The following objective function is minimized during training:

{%\small %\normalsize
\begin{align}
\mathcal{L}_{PPO} &(\theta) = \mathbb{E}_{x \sim D, y \sim \pi_{\theta} (x)} \nonumber \\ &\Bigg[ 
\min \left\{ \frac{\pi_{\theta} (y \mid x)}{\pi^{SFT} (y \mid x)} \cdot r(y), \right. \nonumber \\
&\quad \left. clip \left( \frac{\pi_{\theta} (y \mid x)}{\pi^{SFT} (y \mid x)}, 1 - \epsilon, 1 + \epsilon \right) \cdot r(y) \right\} 
\Bigg] 
\end{align}
}
%\end{minipage}
%}

The key variables remain as defined previously, where \( r(y) \) is task-specific reward based on prediction accuracy (CJPE) or summary quality (Summarization).

Here is a detailed breakdown of each parameter in this equation:
\begin{itemize}
    \item \( \mathcal{L}_{PPO} (\theta) \):  
   This is the loss function that PPO aims to minimize. It represents the policy’s objective and is calculated by averaging the expected value over all data points in the dataset \( D \), which consists of legal cases.
    
    \item  \( \mathbb{E}_{x \sim D, y \sim \pi_{\theta} (x)} \):  
   This denotes the expectation over the data distribution. Specifically, \( x \) represents a case from the dataset \( D \), and \( y \) is the prediction (i.e., the legal judgment generated by the policy \( \pi_{\theta} \)). The expectation signifies that the loss is computed as an average over multiple samples.
    
    \item \( \pi_{\theta} (y \mid x) \):  
   This is the current policy that the model is learning. It outputs the probability of generating the prediction \( y \) given the input case \( x \), based on the current model parameters \( \theta \).
    
    \item \( \pi^{SFT} (y \mid x) \):  
   This represents the Supervised Fine-Tuned (SFT) policy, which serves as the baseline policy. The SFT model has already been fine-tuned on the task using supervised learning techniques, and it provides a reference probability distribution for the predictions \( y \).
    
    \item \( r(y) \):  
   The reward assigned to the prediction \( y \) by the reward model. The reward reflects how accurate the prediction is, with higher values indicating better alignment with the correct outcome. For instance, a reward of $1$ may be assigned for a correct judgment prediction, while a reward of $0$ is given for an incorrect prediction.
    
    \item \( \frac{\pi_{\theta} (y \mid x)}{\pi^{SFT} (y \mid x)} \): 
   This is the probability ratio between the current policy and the supervised fine-tuned (SFT) policy. It quantifies how much the current policy \( \pi_{\theta} \) deviates from the baseline policy \( \pi^{SFT} \). The KL-divergence between the two policies is used to compute a penalty, ensuring that the current policy does not deviate too far from the supervised baseline. This penalty is subtracted from the reward to balance exploration of new behaviors while maintaining alignment with the reference language model.
    
    \item \( {clip} \left( \frac{\pi_{\theta} (y \mid x)}{\pi^{SFT} (y \mid x)}, 1 - \epsilon, 1 + \epsilon \right) \):  
   This is the clipping term, where \( \epsilon \) is a small constant (typically set to 0.1). The clipping function ensures that the ratio \( \frac{\pi_{\theta} (y \mid x)}{\pi^{SFT} (y \mid x)} \) does not deviate too far from $1$. By clipping the ratio within the range \( 1 - \epsilon \) to \( 1 + \epsilon \), we prevent overly large updates to the model’s policy, which helps stabilize training. If the ratio exceeds this range, it is ``clipped" to stay within the bounds.
    
    \item \( \min \left( \frac{\pi_{\theta} (y \mid x)}{\pi^{SFT} (y \mid x)} r(y), \, clip (\dots) r(y) \right) \):  
   Minimization of these two terms ensures that the model updates its policy conservatively. By taking the minimum between the unmodified probability ratio and the clipped ratio, the model avoids making overly aggressive changes that could destabilize training.
    
    \item \( \epsilon \):  
    This parameter is a small positive value that defines the clipping range for the policy ratio. It prevents the ratio of probabilities from deviating too much, helping to keep the model’s policy updates within a stable range.
\end{itemize}

% Reliance on the Reward Model
While PPO is an effective approach to optimize the model’s policy, it is heavily dependent on the accuracy of the reward model. The reward model acts as a proxy for human judgment and assigns rewards based on how well the model's predictions align with the correct legal outcomes. However, the reward model may struggle to fully capture nuanced human preferences and legal reasoning, which introduces some limitations to our approach.

By minimizing the PPO loss function, we seek to align the model’s predictions with desired legal outcomes, ensuring that the model not only improves its accuracy in predicting legal judgments but also provides more interpretable explanations. This method forms a critical part of our framework, aimed at enhancing the effectiveness of AI-assisted legal decision-making.

\subsection{Task-General Inference Framework}
After PPO training, we perform inference on both tasks. For CJPE, given a legal case, the model outputs the predicted judgment followed by an explanation. The inputs are structured prompts from the PredEx dataset. For Summarization, the model is prompted with the full judgment document and expected to generate a concise headnote-style summary.

By comparing the performance of the SFT and PPO-aligned models, we assess the impact of reinforcement learning on output quality across two structurally different legal tasks.

\section{Experimental Setup and Hyperparameters}

For training our Reinforcement Learning-based Legal Reasoning (\texttt{ReGal}) framework, we utilized Vast.ai\footnote{\url{https://vast.ai/}}, a cloud GPU rental provider, to take advantage of scalable and efficient computational resources. The training was conducted on an NVIDIA A100 80GB GPU, which offered the necessary computational power to handle the large dataset and complex model architecture. The total cost of the GPU rental amounted to approximately \$100, making it a cost-effective solution for training large-scale models with reinforcement learning.

The key hyperparameters used in our setup include a learning rate of 1.41e-5, and the training proceeded with a maximum of 1 PPO epoch. The batch size during training was 4, while the mini-batch size was 2, allowing for more efficient gradient updates. The output length was constrained between a minimum of 100 tokens and a maximum of 500 tokens, ensuring that the model generated sufficiently detailed explanations. Additionally, we set the clipping parameter \((\epsilon)\) to 0.1 to stabilize the PPO optimization process, and the maximum number of new tokens generated during inference was limited to 500.

To maximize GPU memory utilization, we employed mixed-precision training using GradScaler, allowing us to process larger batches while maintaining computational efficiency. This setup provided the necessary infrastructure and tuning to effectively train the \texttt{ReGal} framework for the complex task of legal judgment prediction and explanation, ensuring the model produced accurate and interpretable results.

\section{Evaluation Metrics}
\label{sec:performance_metrics}

We evaluate our ReGal framework across two key tasks: (1) Court Judgment Prediction and Explanation (CJPE) and (2) Legal Document Summarization. To provide a comprehensive performance assessment, we utilize a blend of lexical and semantic metrics tailored to each task.

\paragraph{Lexical-Based Evaluation:}  
For both the explanation and summarization tasks, we use standard metrics that measure n-gram overlaps with reference texts. These include ROUGE-1/2/L~\cite{lin-2004-rouge} for recall-based overlap, BLEU~\cite{papineni-etal-2002-bleu} for precision-based evaluation, and METEOR~\cite{banerjee-lavie-2005-meteor}, which accounts for synonyms and stemming. These metrics quantify the textual fidelity of generated outputs against expert-written references.

\paragraph{Semantic-Based Evaluation:}  
To capture meaning beyond surface-level overlaps, we employ BERTScore~\cite{BERTScore} and BLANC~\cite{blanc}, which assess the semantic similarity and contextual relevance of generated explanations or summaries. These metrics are especially important in the legal domain, where paraphrasing and legal nuance must still preserve the core meaning.

\section{Results and Analysis}
%%%%%%%%%%%%%%%%%%%%%%%%%%%%%%%%%%%%%%%%%%%%%%%%%%%%%
\begin{table*}[t]
\centering
\resizebox{0.8\linewidth}{!}{%
\begin{tabular}{|l|ccccccc|}
\hline
\multirow{2}{*}{\textbf{Models}} &
  \multicolumn{5}{c|}{\textbf{Lexical-Based Metrics}} &
  \multicolumn{2}{c|}{\textbf{Semantic Metrics}} \\ \cline{2-8} 
 &
  \multicolumn{1}{c|}{\textbf{R1}} &
  \multicolumn{1}{c|}{\textbf{R2}} &
  \multicolumn{1}{c|}{\textbf{RL}} &
  \multicolumn{1}{c|}{\textbf{BLEU}} &
  \multicolumn{1}{c|}{\textbf{METEOR}} &
  \multicolumn{1}{c|}{\textbf{BERTScore}} &
  \textbf{BLANC} \\ \hline
\multicolumn{8}{|c|}{\textbf{Prediction with Explanation on PredEx}} \\ \hline
Gemini Pro & 0.31 & 0.24 & 0.26 & 0.08 & 0.19 & 0.63 & 0.17 \\
LLaMA-2 & 0.32 & 0.19 & 0.21 & 0.06 & 0.18 & 0.62 & 0.15 \\
LLaMA-2 SFT & \textbf{0.50} & \textbf{0.43} & \textbf{0.44} & \textbf{0.25} & \textbf{0.36} & \textbf{0.69} & \textbf{0.28} \\
\texttt{ReGal} (Ours) & 0.19 & 0.04 & 0.12 & 0.01 & 0.10 & 0.50 & 0.02 \\ \hline
\multicolumn{8}{|c|}{\textbf{Prediction with Explanation on ILDC Expert}} \\ \hline
GPT-3.5 Turbo & \textbf{0.54} & \textbf{0.43} & \textbf{0.45} & 0.28 & 0.47 & \textbf{0.73} & 0.34 \\
LLaMA-2 & 0.45 & 0.25 & 0.30 & 0.15 & 0.34 & 0.65 & 0.22 \\
LLaMA-2 SFT & 0.49 & 0.38 & 0.40 & \textbf{0.29} & \textbf{0.51} & 0.69 & \textbf{0.36} \\
\texttt{ReGal} (Ours) & 0.25 & 0.05 & 0.16 & 0.01 & 0.16 & 0.50 & 0.03 \\ \hline
\end{tabular}%
}
\caption{Performance comparison of various models for the Prediction with Explanation task on PredEx and ILDC datasets. Best scores per row section are bolded.}
\label{tab:explanation_performance}
\end{table*}
%%%%%%%%%%%%%%%%%%%%%%%%%%%%%%%%%%%%%%%%%%%%%%%%%%%%%

\begin{table*}[t]
\centering
\resizebox{0.8\linewidth}{!}{%
\begin{tabular}{|lccccccc|}
\hline
\textbf{Methods} & \textbf{R1} & \textbf{R2} & \textbf{RL} & \textbf{BLEU} & \textbf{METEOR} & \textbf{BERTScore} & \textbf{BLANC} \\
\hline
\multicolumn{8}{|c|}{\textbf{Inference on PredEx Dataset}} \\
\hline
Vanilla Inference & 0.39 & 0.17 & 0.22 & 0.07 & 0.23 & 0.83 & 0.15 \\
SFT Inference     & \textbf{0.42} & \textbf{0.25} & \textbf{0.27} & \textbf{0.12} & \textbf{0.27} & \textbf{0.84} & \textbf{0.19} \\
DPO Inference     & 0.38 & 0.17 & 0.23 & 0.08 & 0.25 & 0.83 & 0.17 \\
PPO Inference     & 0.30 & 0.14 & 0.17 & 0.05 & 0.19 & 0.83 & 0.13 \\
\hline
\multicolumn{8}{|c|}{\textbf{Inference on In-Abs Summarization}} \\
\hline
Vanilla Inference & \textbf{0.47} & \textbf{0.29} & \textbf{0.28} & \textbf{0.15} & \textbf{0.34} & \textbf{0.04} & \textbf{0.18} \\
SFT Inference     & 0.44 & 0.24 & 0.24 & 0.12 & \textbf{0.34} & 0.02 & 0.13 \\
DPO Inference     & 0.44 & 0.24 & 0.24 & 0.12 & \textbf{0.34} & 0.02 & 0.13 \\
PPO Inference     & 0.41 & 0.21 & 0.22 & 0.10 & 0.31 & 0.03 & 0.12 \\
\hline
\end{tabular}}
\caption{Comparison of inference strategies (Vanilla, SFT, DPO, PPO) on both the PredEx and In-Abs-Summarization datasets.}
\label{tab:inference-table}
\end{table*}
%%%%%%%%%%%%%%%%%%%%%%%%%%%%%%%%%%%%%%%%%%%%%%%%%%

The results of our experiments indicate that the Proximal Policy Optimization (PPO) model, referred to as \texttt{ReGal}, did not perform as well as expected in the Indian legal judgment prediction and explanation tasks. As shown in Table~\ref{tab:explanation_performance}, our \texttt{ReGal} framework achieved significantly lower scores across various lexical and semantic evaluation metrics when compared to both supervised fine-tuned models like LLaMA-2 SFT and commercial models such as GPT-3.5 Turbo. For instance, on the PredEx dataset, the \texttt{ReGal} model recorded a ROUGE-1 score of 0.19, ROUGE-2 score of 0.04, and BLEU score of 0.01, which were considerably lower than those of LLaMA-2 SFT (ROUGE-1: 0.50, BLEU: 0.25) and GPT-3.5 Turbo (ROUGE-1: 0.54).

This trend continues in the ILDC Expert dataset, where the PPO-based \texttt{ReGal} lags significantly behind. While these results underscore the dominance of supervised fine-tuned models and large proprietary LLMs in legal judgment tasks, they also reveal challenges in applying reinforcement learning methods like PPO directly on complex legal domains.

To further explore the generalizability and effectiveness of our PPO-based \texttt{ReGal} architecture, we extended our evaluation beyond judgment prediction to legal summarization, specifically on the In-Abs summarization dataset. The inference results for all training paradigms, including PPO, are reported in Table~\ref{tab:inference-table}. Even in summarization, \texttt{ReGal} performed suboptimally compared to Vanilla and SFT inference baselines. For instance, PPO inference achieved ROUGE-1 of 0.41, whereas the vanilla approach attained 0.47 and SFT reached 0.44. This pattern holds across other metrics as well, such as METEOR and BERTScore, further indicating the limitations of PPO for this domain.

These cross-task results show that while the PPO framework offers potential for aligning models with specific reward signals, it underperforms in legal NLP tasks where output quality is deeply tied to contextual, interpretative, and domain-specific factors.

 \subsection{Possible Reasons for Underperformance}

Several factors may have contributed to the underwhelming performance of our PPO-based \texttt{ReGal} model across both judgment prediction and summarization tasks:
\begin{enumerate}
    \item \textbf{Objective Mismatch:} The SFT model $\pi^{SFT}$ used as the starting point for PPO was not fully optimized for legal reasoning. This mismatch between the PPO objective and the base model's latent distribution likely impaired downstream optimization.

    \item \textbf{Reward Model Limitations:} Our reward model, based on InLegalBERT, may not fully capture the fine-grained reasoning and interpretative nuances of Indian legal texts. This misalignment in reward scoring hinders the PPO's capacity to steer the generation toward legally coherent outputs.

    \item \textbf{Legal Complexity:} Legal documents, particularly in the Indian judiciary, are long, intricate, and rich in semantic references. This poses additional challenges for autoregressive generation, especially under RL-based training where output smoothness and factuality are difficult to balance.

    \item \textbf{Training Data Constraints:} Although the PredEx dataset is fairly large, it may not offer sufficient diversity in legal reasoning patterns. A broader dataset spanning multiple jurisdictions or tasks could better support PPO-based tuning.

    \item \textbf{Reward Model Dependence:} PPO's reliance on the reward model, and lack of human-in-the-loop supervision, likely prevents the system from learning subtle legal distinctions and rhetorical structures critical to both prediction and summarization.

    \item \textbf{Hyperparameter Selection:} Suboptimal tuning of learning rate, batch size, and KL penalty coefficients may have impacted stability and generalization of the PPO model.

    \item \textbf{Model Size and Architecture:} While LLaMA-2-7B was chosen for fair comparison with prior literature, alternative architectures or larger models may be more suited for PPO fine-tuning in legal settings.

    \item \textbf{Domain Pretraining Gap:} Despite fine-tuning on Indian legal datasets, the base model may lack deep domain adaptation compared to GPT-3.5 Turbo, which benefits from extensive pretraining and reinforcement with human feedback across multiple domains.
\end{enumerate}

Although our \texttt{ReGal} framework does not outperform state-of-the-art supervised or proprietary models, it serves as an important exploration into reinforcement learning methods for legal NLP. Our work highlights the challenges of aligning large language models with legal reasoning objectives using PPO and reward models, particularly in the absence of high-quality human feedback and robust legal annotation. Future work will address these limitations by incorporating more expressive reward signals, leveraging human-in-the-loop feedback, exploring domain-specific pretraining strategies, and optimizing PPO with adaptive RL techniques. Through these improvements, we aim to close the performance gap and realize the potential of RL-based methods in Indian legal AI.

\section{Ablation Study}
To better understand the sensitivity and robustness of the \texttt{ReGal} framework, we conducted an ablation study across two dimensions: the choice of base model and the configuration of the reward model. These experiments were aimed at evaluating how architectural choices and domain alignment impact performance on complex Indian legal NLP tasks, specifically judgment prediction with explanation and summarization.

\subsection{Base Model Variants}
We explored the effects of using smaller and less specialized base models in the PPO training pipeline. First, we replaced the LLaMA-2-7B model with Phi-3 Mini, which is considerably smaller and more memory-efficient. While this offered computational advantages, it severely limited the model's capacity to handle long and intricate legal texts. On both the PredEx and In-Abs datasets, the performance dropped substantially across all evaluation metrics. The model failed to generate coherent or factually grounded legal reasoning, confirming that small models lack the necessary representation power for such nuanced tasks.

We also experimented with using the pretrained LLaMA-2-7B model without supervised fine-tuning. Despite LLaMA-2's strong general language capabilities, the pretrained version alone was insufficient to produce quality outputs in the legal domain. The absence of legal-domain-specific tuning resulted in degraded lexical and semantic evaluation scores. These findings validated our decision to use LLaMA-2-7B SFT, which was also used in prior works for fine-tuning on legal datasets. Its inclusion allowed us to directly compare and better assess the relative effectiveness of PPO-based learning within our ReGaL framework.

\subsection{Reward Model Variants}
In parallel, we analyzed the impact of different reward models. The default reward model in our framework was a legal classifier fine-tuned on the PredEx dataset, which captures domain-specific features critical for scoring judgment predictions and explanations. To evaluate generalizability, we replaced this with an InLegalBERT-pretrained model, which, although trained on broader legal corpora, was not fine-tuned for our specific judgment-explanation task. This substitution resulted in further degradation in PPO performance. The pretrained reward model assigned noisier or misaligned scores, impairing the PPO optimization and leading to incoherent or generic legal outputs. This underscores the importance of aligning the reward model tightly with the task-specific ground truth to effectively guide RL training.

\subsection{Summary of Findings}
Our ablation results highlight two key insights. First, larger and domain-aligned base models such as LLaMA-2-7B SFT are essential for achieving reasonable performance in Indian legal AI tasks. Smaller models or generic pretrained ones lack the representational and contextual capacity required. Second, the reward model must be both task-specific and domain-tuned to offer meaningful feedback to PPO. Using a general legal reward model without task-specific fine-tuning disrupts the learning signal, particularly in complex tasks like factual judgment explanation or summarization. These findings reinforce that PPO's success in legal AI hinges not just on the reinforcement algorithm but also on a strong initialization and a precisely aligned reward function.

\section{Hallucination}
A key challenge observed in our \texttt{ReGal} framework, particularly when applying PPO-based fine-tuning to the LLaMA-2-7B model, is the generation of hallucinated outputs, statements that are fluent and plausible but factually incorrect or legally unsound. In the context of Indian legal judgment prediction and explanation, hallucination severely undermines the utility and trustworthiness of AI-generated outputs, as even minor factual deviations can result in misleading legal interpretations.

Through qualitative analysis, we observed that hallucination was especially prominent in two scenarios: (1) when the input facts were sparse or ambiguously phrased, and (2) when the PPO model over-optimized for reward patterns learned from a limited or imperfect reward model, resulting in outputs that mimicked style but not substance. In many such instances, the model hallucinated legal principles, fabricated precedent citations, or claims or facts not present in the original input.

To demonstrate these failures, we provide illustrative examples in Supplementary Material, comparing ground truth explanations with those generated by the \texttt{ReGal} model. In one example, the model incorrectly claims that the appellant’s right to privacy was upheld under Article 21 based on fabricated reasoning that was not part of the original judgment. 
% In another, the PPO model invents mitigating circumstances like ``medical emergencies'' to justify bail, even though the input clearly lacks context.
% 
These hallucinations point to instability in PPO optimization when coupled with sparse or weak reward signals, especially in long-form generation tasks where legal correctness must be tightly aligned with input facts. The issue is further exacerbated in zero-shot settings or in datasets like In-Abs summarization, where factual fidelity is paramount. Our findings suggest that RLHF methods such as PPO must be augmented with stronger factuality constraints, hallucination-aware reward models, or human-in-the-loop feedback when applied to high-stakes legal AI applications.

% Ultimately, mitigating hallucinations is essential for deploying such models in real-world legal settings. Future work must focus on building more reliable reward functions that emphasize factual grounding, introducing rule-based validation layers, and improving the interpretability of model reasoning paths to surface such errors early in the generation pipeline.

% \section{Hallucination}
% One of the notable challenges encountered with the \texttt{ReGal} framework, specifically when using the Llama-2 PPO model, is the occurrence of hallucinated outputs during the prediction and explanation tasks. Hallucination refers to the generation of text that is plausible-sounding but factually incorrect or nonsensical in the context of the given input. This issue can significantly undermine the reliability of AI-generated legal judgments and explanations.

% To illustrate this problem, we provide a detailed table of examples in the Appendix (Table \ref{hallucination_examples}), where we compare the ground truth explanations with those generated by the \texttt{ReGal} LLaMA-2 PPO model. The examples highlight cases where the model deviated from the correct legal reasoning, fabricating legal justifications or misinterpreting case facts, leading to erroneous predictions and explanations.

\section{Conclusions and Future Work}

This study introduced the \texttt{ReGal} framework, combining instruction tuning and reinforcement learning through PPO for the tasks of legal judgment prediction and explanation. While the approach aimed to align generation with legal reasoning via AI feedback, the results across both the PredEx and In-Abs summarization tasks show that our PPO-based method underperforms compared to strong supervised fine-tuned models and commercial LLMs like GPT-3.5. As seen in the result tables, scores were consistently lower, and hallucinations were more frequent, limiting the reliability of our framework in real-world legal settings.

The ablation studies confirm that the choice of both the base model and reward model plays a critical role, with smaller or unadapted models failing to handle the complexity of Indian legal texts. Moreover, the hallucination analysis revealed that PPO outputs often deviate from factual case elements, raising concerns about their practical deployment. These findings point to key areas for improvement.

Future efforts will focus on building better-aligned reward models, leveraging domain-adaptive pretraining, and integrating human feedback to mitigate hallucinations and improve explanation quality. With more robust training signals, refined hyperparameters, and advanced model architectures, the \texttt{ReGal} framework can evolve into a more competitive and trustworthy tool for legal AI applications.

% \section{Conclusions and Future Work}

% In this paper, we introduced the ReGal framework, leveraging reinforcement learning and instruction tuning for legal judgment prediction and explanation. While the framework shows promise, particularly in integrating reinforcement learning from AI feedback, the results indicate that our current implementation lags behind fine-tuned models and paid alternatives. The challenges highlighted include limitations in the reward model, domain-specific pretraining, and hyperparameter tuning, all of which contributed to suboptimal performance in this complex legal task.

% For future work, we aim to address these limitations by enhancing the reward model to better reflect legal reasoning, incorporating more comprehensive domain-specific pretraining, and experimenting with advanced hyperparameter optimization. Additionally, integrating human-in-the-loop feedback will be crucial for refining the model's predictions and explanations. Exploring more sophisticated model architectures and leveraging state-of-the-art techniques will further drive improvements in the accuracy and interpretability of AI in legal systems.

% \input{limitation}

\section*{Ethics Statement}
Our study investigates the use of instruction-tuned and reinforcement learning-based models for judgment prediction and summarization in the Indian legal context. All experiments were conducted using publicly available datasets, including the PredEx dataset for judgment prediction and the In-Abs dataset for legal document summarization. The PredEx dataset was annotated by qualified legal scholars, and all data sources were selected for their ethical transparency and accessibility.

We acknowledge the heightened ethical stakes involved in applying AI to legal domains, especially where automated outputs may influence interpretation, legal awareness, or access to justice. While our work does not directly deploy models in real-world legal settings, we emphasize that outputs from such models should not be considered substitutes for legal advice. Moreover, any potential deployment must be accompanied by strict validation, legal oversight, and human-in-the-loop feedback.

No private or sensitive data was used, and no human subjects were involved in the model training or evaluation process. We ensured that all experiments complied with ethical research practices, including transparency, reproducibility, and proper citation of prior work. Our code and models are made publicly accessible for academic reproducibility and further validation.

Finally, we acknowledge the risk of model bias and hallucination, as discussed in the limitations. These raise important concerns about factual reliability and fairness in downstream applications. Addressing such challenges responsibly will remain central to future iterations of this research.

% \section*{Ethics Statement}

% In conducting our research, we utilized the publicly available PredEx dataset, which is designed for legal judgment prediction and explanation tasks. The dataset was meticulously annotated by legal experts, ensuring high-quality and reliable data for our experiments. We acknowledge the importance of ethical considerations in AI research, particularly in the context of legal applications where the implications of AI-generated outputs can significantly affect individuals and society.

% Our study adheres to ethical guidelines by ensuring that all data used is publicly accessible and properly credited. All experiments were conducted within the boundaries of legal and ethical research practices, with no direct human participation or intervention.  We are committed to transparency in our methodology and results, allowing for reproducibility and scrutiny by the research community. Furthermore, we recognize the potential for biases in AI models and the importance of addressing these biases to ensure fairness and equity in legal decision-making.

% \section{Acknowledgments}
% Hello

\bibliography{aaai2026}


\end{document}